\documentclass[letterpaper, 10 pt, conference]{IEEEtran}
\IEEEoverridecommandlockouts
\usepackage{fancyhdr}
\usepackage[utf8]{inputenc}
\usepackage[T1]{fontenc}
\usepackage{cite}
\makeatletter
\let\NAT@parse\undefined
\makeatother
\usepackage[colorlinks,citecolor=black, linkcolor=black]{hyperref}
\usepackage{graphics}
\usepackage{epsfig}
\usepackage{mathptmx}
\usepackage{mathptmx}
\usepackage{amsmath}
\usepackage{amssymb}
\usepackage{makecell}

\def\R{\mathbb{R}}

\def\CEN{\mathrm{CEN}}

\title{\LARGE \bf SPFCN: Select and Prune the Fully Convolutional Networks for Real-time Parking Slot Detection}
\author{
 Zhuoping Yu$^{1}$, Zhong Gao$^{1}$, Hansheng Chen$^{1}$, Yuyao Huang$^{1, 2, *}$
 \thanks{$^{1}$ Institute of Intelligent Vehicles, School of Automotive Studies, Tongji University, Shanghai, China}
 \thanks{$^{2}$ State Key Laboratory of Advanced Design and Manufacturing for Vehicle Body, China}
 \thanks{$^{*}$ Corresponding author. E-mail: huangyuyao@tongji.edu.cn}
 \thanks{The model will be posted on \url{https://github.com/tjiiv-cprg/SPFCN-ParkingSlotDetection}}
}

\fancypagestyle{FirstPage}{
\fancyhead[L]{
  \copyright~2020 IEEE. Personal use of this material is permitted. Permission from IEEE must be obtained for all other uses, in any current or future media, including reprinting/republishing this material for advertising or promotional purpose, creating new collective works, for resale or redistribution to servers or lists, or reuse of any copyrighted component of this work.}
\fancyfoot[]{}
}

\begin{document}

\maketitle
\thispagestyle{FirstPage}


\begin{abstract}

For vehicles equipped with the automatic parking system, the accuracy and speed of the parking slot detection are crucial. But the high accuracy is obtained at the price of low speed or expensive computation equipment, which are sensitive for many car manufacturers. In this paper, we proposed a detector using CNN(convolutional neural networks) for faster speed and smaller model size while keeps  accuracy. To achieve the optimal balance, we developed a strategy to select the best receptive fields and prune the redundant channels automatically after each training epoch. The proposed model is capable of jointly detecting corners and line features of parking slots while running efficiently in real time on average processors. The model has a frame rate of about 30 FPS on a  2.3 GHz CPU core, yielding parking slot corner localization error of 1.51$\pm$2.14 cm (std. err.) and slot detection accuracy of 98\%, generally satisfying the requirements in both speed and accuracy on on-board mobile terminals.

\end{abstract}


\section{INTRODUCTION}
With the development of autonomous driving technologies, automatic parking assist system has become an intensive research topic and its first step is to find the parking slot. The proposed solutions can be roughly divided into two categories: one is infrastructure based approach which typically use pre-built maps and sensors, and the other is on-board sensor based approach which use only sensors mounted on vehicles to detect available parking slots. In this paper, we used second approach to detect the parking slot makings which are group of line segments drawn on the ground to indicate an effective parking area.


\begin{figure}[tbp]
 \centering
 \includegraphics[width=75mm]{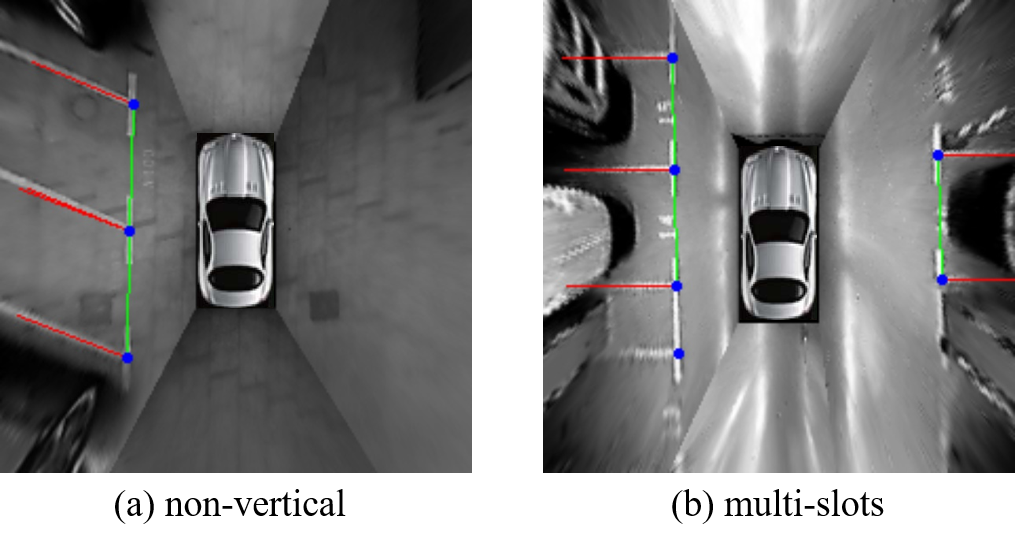}
 \caption{Detection results with corners marked as blue points, entry lines marked green and separating lines marked red. The synthetic overhead view is composed by images from four calibrated, undistorted and homographic transformed fish-eye cameras mounted on the vehicle, according to \cite{zhang_vision-based_2018}. The artifacts and misalignments aused by calibration error can be further reduced by additional transformation \cite{Bruls_Porav_Kunze_Newman_2019} and stitching algorithms, but we only focus on the detection pipeline in this paper.}
 \label{Model Result abstract}
\end{figure}


The detection methods based on locating visual markers can again be divided into two categories: line-based~\cite{wu_vh-hfcn_2018} and corner-based~\cite{zhang_vision-based_2018-1}. In this paper, a pure visual method is designed to find the parking slots, in which both line features and corner features will be detected to improve accuracy. Our method differs from other parking slot detectors in three major contributions as concluded below:

\begin{itemize}
 \item First, we introduce the hourglass structure~\cite{newell_stacked_2016} into parking slot detection and discuss and verify the effect of  the network with/without stacked hourglass and intermediate supervisions.
 \item Second, we design a Select-and-Prune Module (SP Module), which can select the most suitable receptive field for convolution, and prune redundant channels automatically. The module can be easily used in training epoch and will not bring extra calculations to the inference stage.
 \item Third, we use both corner features and line features to jointly infer the parking slot. The joint features bring better stability in inferencing and can maintain a certain accuracy when continuously compressing the model parameters.
\end{itemize}


\section{RELATED WORKS}
\textbf{Parking Slot Detection: } One of the first works to use traditional machine learning method for parking slot detection was presented by Xu et al.~\cite{jin_xu_vision-guided_2000}. Later implementations of traditional methods include using Radon transform~\cite{wang_automatic_2014} to detect slot lines or using Harris detector~\cite{suhr_sensor_2014}, AdaBoost~\cite{zhang_vision-based_2018} to detect slot corners. The results of these methods largely depend on the feature design of classifiers and subsequent logical inference, and they are also sensitive to the size or direction of input images. The occluded and shadowed ground markings also bring extra difficulties in solving the detection problem with traditional methods.

As for the deep-learning based methods, for example, Zhang et al.~\cite{zhang_vision-based_2018-1} use YOLO v2~\cite{redmon_yolo9000:_2016} to locate parking slot corners. However, YOLO is a object detection network designed not exclusively for parking slot and running YOLO in real time on CPU can be a serious problem due to its massive number of FLOPs. Today's production cars are often fitted with only CPUs or even MCUs. Hence, there still lies potential in the performance of the backbone itself.

\textbf{Hourglass-like Network Structure: } In this paper, the hourglass structure~\cite{newell_stacked_2016} is selected as our backbone. This architecture is specifically designed for key-point detection and has been widely used in many related tasks such as the human joint detection.

\begin{figure}[tbp]
 \centering
 \includegraphics[width=50mm]{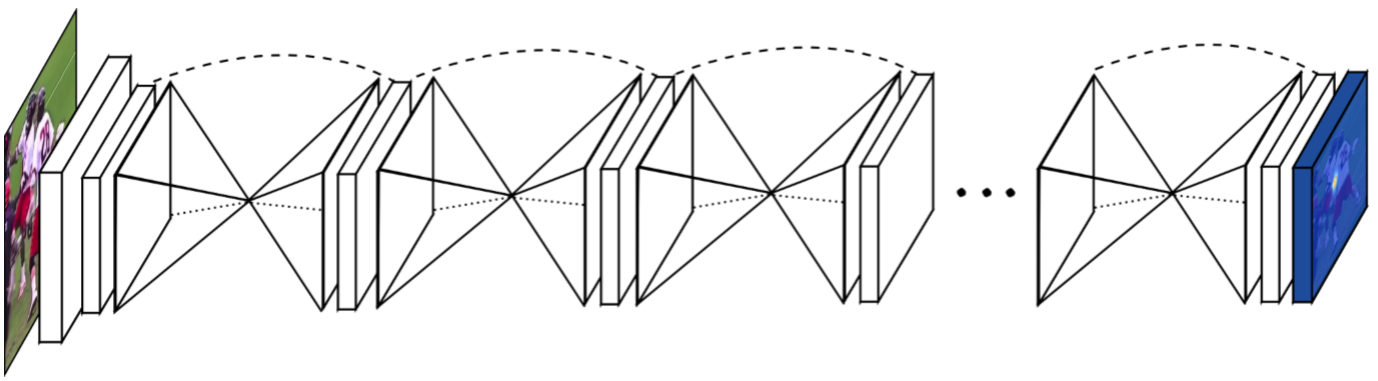}
 \caption{Stacked hourglass network~\cite{newell_stacked_2016}.}
 \label{Stacked Hourglass Network}
\end{figure}

The architecture is shown in Fig.~\ref{Stacked Hourglass Network}. It has the flexibility in stacking the hourglasses to adjust the number of parameters and the speed of network. The input image is first down-sampled, the possible regions of corner-points and lines are found on the smaller feature map, and the feature size is restored by up-sampling, where key areas detected by the previous layer are superimposed on the features of skip, thereby it can increase the response of key areas on the final heat map.

\textbf{Efficient Convolution Module Design: } Regarding basic convolution modules, there is an increasing need for the efficient design. SqueezeNet~\cite{iandola_squeezenet:_2016} implemented AlexNet-level accuracy on ImageNet with a 50-fold reduction in parameters. The MobileNet~\cite{howard_searching_2019} and ShuffleNet~\cite{ma_shufflenet_2018} use depth-wise separable convolution to compress parameters and FLOPs of the network while ensuring accuracy. Based on ResNet~\cite{he_deep_2015} and DenseNet~\cite{huang_densely_2016}, PeleeNet~\cite{wang_pelee:_2018} uses a two-way dense layer and a stem block to optimize its performance.

Li et al.~\cite{li_selective_2019} realized that the convolution kernel size of visual neurons is rarely considered when constructing the CNN. They propose a dynamic selection mechanism to adaptively adjust the weights of kernels in different sizes when fusing multiple branches containing information of different spatial scales.

Inspired by that, we design a new convolution module which can adaptively select the convolution kernel size and prune the redundant channels. We call it the SP Module and substitute it for the residual modules~\cite{he_deep_2015} in the original stacked hourglass network.


\section{METHOD}

We developed an efficient and practical solution for simultaneous detection of corners and marking lines of parking slots. The general idea is using FCN (Fully Convolutional Networks) to accomplish all the jobs, without the help of FC (Fully Connected) layers. The classification, regression, detection, and segmentation tasks generally can be modeled as a softmax heatmap generator followed by a certain type of proposal (e.g. Region Proposal Network~\cite{ren_faster_2015}, etc.), pooling (e.g. ROI Pooling~\cite{ren_faster_2015}, Line Pooling~\cite{Lee_Kim_LinePooling}, Average Pooling), and finally thresholding.
We then develop a scalable and tailorable training strategy for pre-training the model, selecting from the candidate kernels while training, and pruning selected kernels' channels while fine-tuning.
The trained networks with simple post-processing logic can show good efficiency and flexibility to be generalized to other problems by following the similar solution paradigm, the solution for parking slot detection in this paper is an example of such idea.

\subsection{Network Structure}
\begin{figure}[tbp]
 \centering
 \includegraphics[width=60mm]{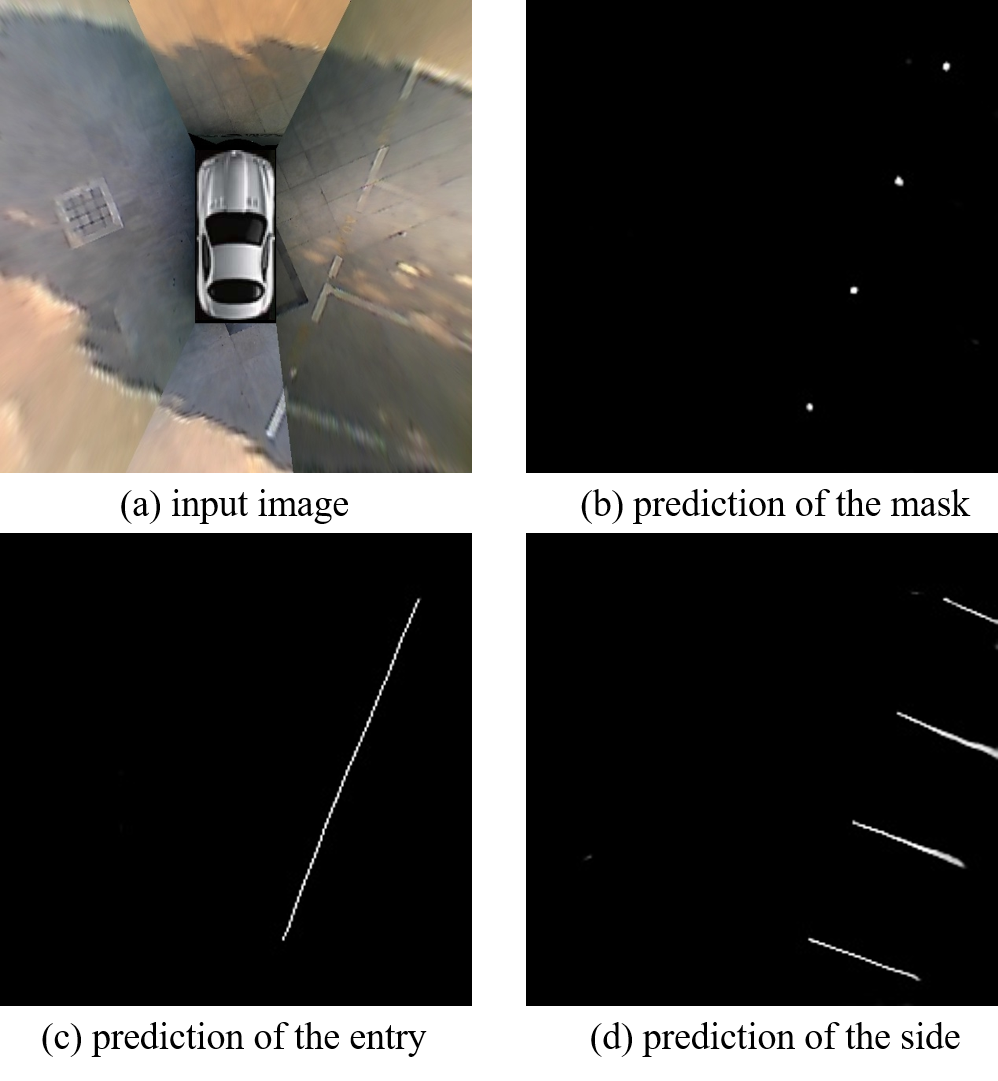}
 \caption{Predicted heatmaps of the Heads.}
 \label{Prediction Heads}
\end{figure}

Among all the FCNs, we choose the Stacked Hourglass Network~\cite{newell_stacked_2016} as the basic architecture, for its scalability to different layers of stacks. The prediction heads of the parking slot detection task produces heatmaps of slots' corners, entrance lines and separating lines, as shown in Fig.~\ref{Prediction Heads}. The final layers, as well as the intermediate supervision layers, supervise the layers by softmax losses through those heads. The existence of the intermediate supervision layers can assist the back propagation of the loss.

\subsection{SP Module}
We propose the Select-Prune Module, which is able to automatically SELECT the best convolution kernel candidates, and PRUNE the least contributional channels in the selected kernel. It works as a substitute for residual block~\cite{he_deep_2015} in the original stacked hourglass architecture.

\textbf{Select Module: } As we know, only kernels with "correct" receptive field can respond to the "correct" feature, making the receptive field a critical hyper-parameter to tune. Thanks to dilated convolution, using multiple receptive fields in one convolutional layer can be done without adding too much computation by composing convolutional kernels with different dilation values. However, a straightforward combination of all different dilation values could be wasteful on mobile devices. Instead, an effort to hunt for a most efficient combination in different layers need to be made. The Select Module is designed for such a purpose, to select the convolution kernel with the best receptive fields in a convolutional layer.

The Select Module has several candidate convolution kernels with different dilation values and its task is to select the best one among them. Input features are fed into each kernel and the contribution is evaluated by a MLP block called \textit{Contribution Evaluation Networks (CEN)}, and outcome of all candidate kernels are weighted by the predicted contribution of the CENs and then summed together as the final outcome of the Select Module. The CENs and the redundant convolution kernel are only involved during the training process, while during inference only the kernel with most average contribution is kept. In other words, when the optimal dilation value of the Select Module is determined, the module will degenerate into an ordinary dilation convolution, and become easy to implement using any frameworks.

\begin{figure}[tbp]
 \centering
 \includegraphics[width=80mm]{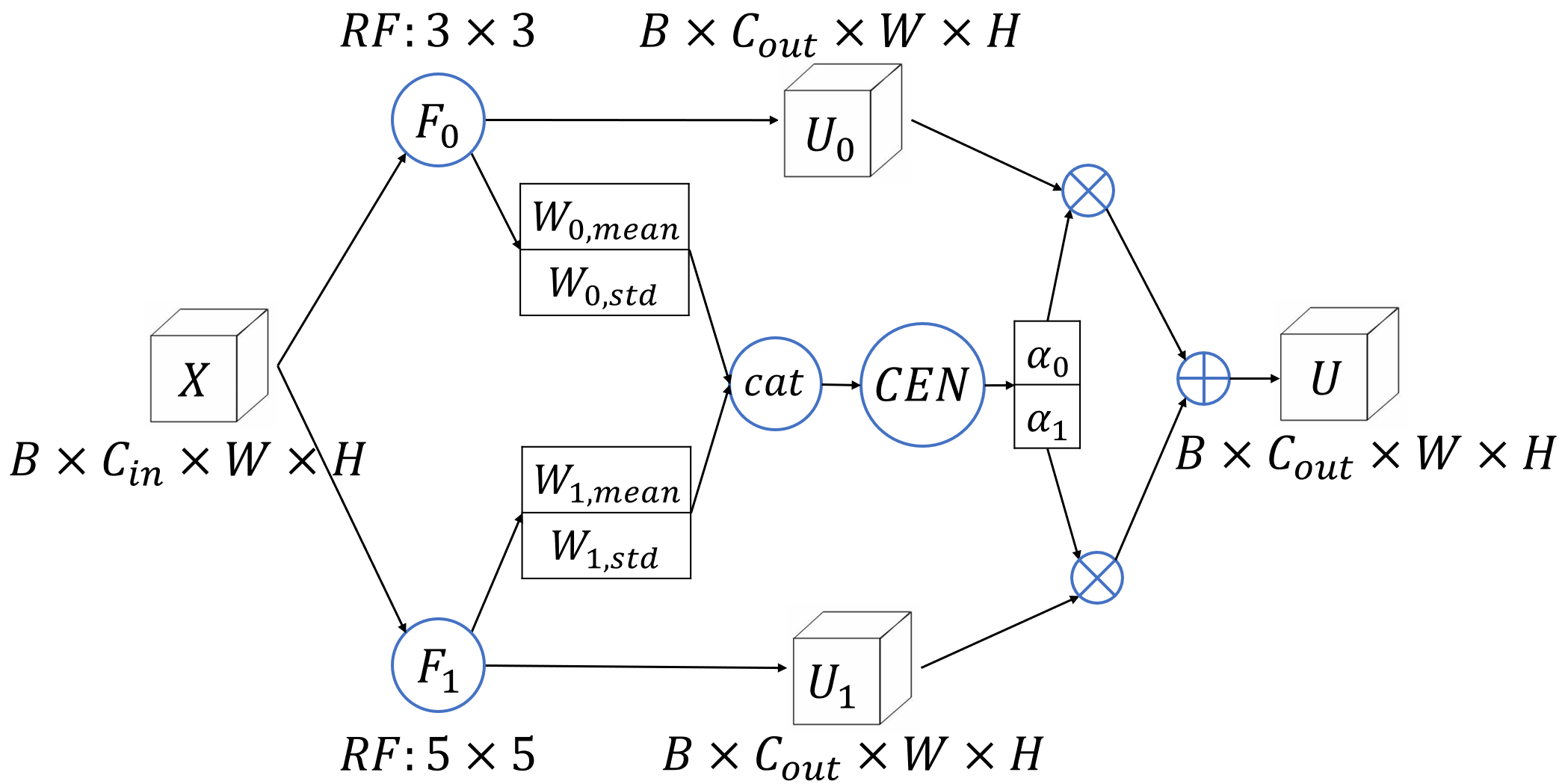}
 \caption{The structure of Select Module. $B$: batch size, $C_{in}$ and $C_{out}$: number of the input/output feature maps' channels, $W$ and $H$: the size of simple feature map, $RF$: receptive field.}
 \label{Structure diagram of Select Mode}
\end{figure}

Here we give an example of the Select Module with capacity 2 in Fig.~\ref{Structure diagram of Select Mode}. For an input feature $X\in\R^{B\times C_{in}\times W\times H}$, we denote the two convolution kernels as $F_0:X\to U_0\in \R^{B\times C_{out}\times W\times H}$ and $F_1:X\to U_1\in \R^{B\times C_{out}\times W\times H}$, with the dilation value of 1 and 2 respectively, which means they are just like a basic convolutional layer with a kernel size of 3 and 5 in terms of receptive fields. The input of the $l^{th}$ convolutional layer's CEN is the flattened parameter of all candidate kernels, denoted as $W_{i,mean}$ and $W_{i,std}$.

Although in the example there are only two kernel candidates, the Select Module can be easily applied to the case with multiple dilation values. We have:
\begin{equation}
	\begin{aligned}
		Vector[\alpha^l] &= \CEN(Vector[W^l]) \\ &= Softmax(FC(ReLU(FC(Vector[W^l]))))
	\end{aligned}
\end{equation}
and
\begin{equation}
 U_{output}^l = \sum_i{\alpha_i^l U_{i}^l},
\end{equation}
in which $\alpha_i^l$ is the summing weights of each kernel's output tensor of the $l^{th}$ layer.

In order to highlight the optimal kernel, we need to obtain sparse weights for different kernels. Considering that $\sum_i^l{\alpha_i^l} = 1$ because of the softmax, we designed the following regularization term to do that:

\begin{equation}
 L_{\CEN}^l = -log(\sum_{i}{{\alpha_i^l}^2})
\end{equation}
because
\begin{equation}
 \sum_{i}{{\alpha_i^l}^2} \le \sum_{i}{{\alpha_i^l}^2}+\sum_{i,j}{\alpha_i^l\alpha_j^l} = (\sum_{i}{\alpha_i^l})^2 = 1
\end{equation}
and only when one term $\alpha_i^l$ equals one and others become zero, $\sum_{i}{{\alpha_i^l}^2}$ can reach the maximum value 1 and $L_{\CEN}^l$ obtains its minimum value 0. With such regularization, vector $[\alpha^l]$ tends to be sparser as driven by the loss function.

\textbf{Prune Module: }
The network is being trained and pruned altenatively~\cite{molchanov_pruning_2016}, splitting the learning procedure into interleaved training phases and pruning phases. The network may go through several epochs during one training phase, and during the following pruning phase, the candidate convolutional channels are traversed, and those that contribute less than a certain level are pruned. The contribution score is calculated by the norm of the weights, as in~\cite{li_pruning_2016}. In case of that all the channels play reasonable roles and the network stops pruning too early, we decide to constantly prune channels of the "global" minimum contribution among all layers in the network each time, thus the network will continue pruning to achieve every configuration of the trade-offs. To make contributions in different layers comparable, we use softmax to normalize the contribution of different channels in the same layer before compared with channels from other layers.

In short, the contribution of the $i^th$ channel in $l^{th}$ layer is calculated as:
\begin{equation}
 contribution_i^l = \frac{exp(\|W_i^l\|_2)}{\sum_j^l{exp(\|W_j^l\|_2)}}
 \label{contribution}
\end{equation}
and there are two types of channel that should be pruned:

\begin{itemize}
\label{pruning_str}
 \item The channel's contribution is too small in its layer, e.g., less than 1\%;
 \item The channel's contribution is one of the smallest 5 in the whole model.
\end{itemize}

\begin{figure}[tbp]
 \centering
 \includegraphics[width=75mm]{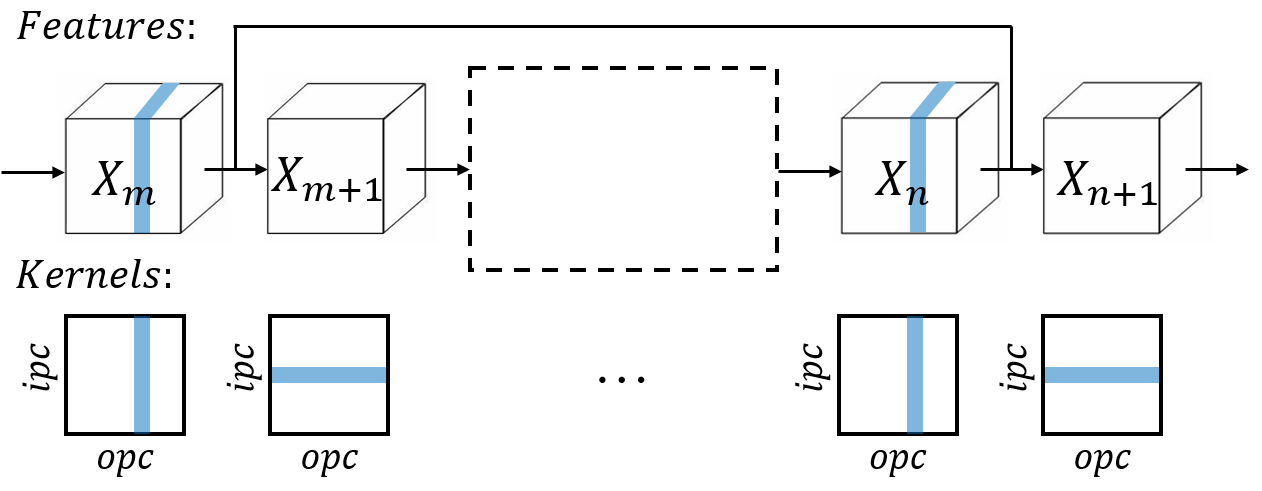}
 \caption{Pruning in hourglass, blue parts represent pruned channels.}
 \label{Pruning in hourglass}
\end{figure}

For a simple convolutional network, using the above pruning strategy is sufficient. But since the hourglass architecture used in this paper has skip connections, special considerations are needed to keep the manifold consistent. The approach is shown in Fig.~\ref{Pruning in hourglass}.

Pruning the $n$-th convolution output affects not only the input of the $(n+1)$-th layer, but also the input and output channels of the layer from skip connections. Like the method mentioned in~\cite{you_gate_2019}, the convolution kernels connected by skip connections are grouped into the same group, and the score is the sum of the pruning scores of each convolution kernel in the group.

Similarly, we use the $L1$ norm of each channel's weight to make them sparse in order to prune the unimportant channels.

\subsection{Slots Inference}
The network's input is a bird eye view (BEV) image, and the output are estimated slot corners and marking lines (including entry lines and separating lines).

The surround-view image sequence could be synthesized in real-time during inference from outputs of multiple wide-angle cameras mounted on the vehicle. Four fish-eye cameras are used, and the surround-view is the composite view of them: the front, left, rear and right view. A lookup table is used to calculate the relations between the fish-eye image and the BEV. The transformation matrix from the BEV image coordinate system to the world coordinate system originating at the vehicle center is calibrated beforehand. Once a parking slot is detected in the BEV, its world coordinates can be calculated. Further details about BEV image synthesis can be found in~\cite{zhang_vision-based_2018}.

We first detect the corners of slots by finding local maximums on the corner heatmap. Then, corners are paired and screened with several rules to decide whether the two corners belong to one slot. Many prior and posterior knowledge can serve as the rules. For example, entrance corners should be within reasonable distances, and no third corner points should intercept between them; standard parking slots form angles of $90^\circ$, $45^\circ$ or $135^\circ$ between entrance line and separating lines, and the average scores along proposed separating lines on those direction can be verified for testing the existence of the lines. The filtered lines and corners are finally assembled together as the detected parking slots.

\section{EXPERIMENTS}
\subsection{Dataset}
We train and test on the challenging DeepPS~\cite{zhang_vision-based_2018} dataset, which includes various scenarios, weather, time and parking slot types. There are 9527 training images and 2138 validating BEV images, and the dataset can be found at \url{https://cslinzhang.github.io/deepps/}.

\subsection{Metrics}
Here we follow the definition of Zhang et al.~\cite{zhang_vision-based_2018-1}:
for a ground-truth marking-point $g_i$, if there is a detected marking point $d_i$ satisfying $e_i=|g_i-d_i|\le\delta$, where $e_i$ is the localization error and $\delta$ is a predefined threshold, we deem that $g_i$ is correctly detected and $d_i$ is a true positive. In this paper, we evaluate the performance of the model by the \textit{accuracy} of slot corner detection with the tolerance of 6 cm by default (approximately 1.5 pixels in the image), instead of 16 cm which was set by Zhang et al.~\cite{zhang_vision-based_2018}~\cite{zhang_vision-based_2018-1}.

\subsection{Training Details}
The input images for training are 224$\times$224 BEV gray-scale images, with corner points and slot line labels. The whole training process can be divided into three stages, including a pre-training stage, a selecting stage, and a pruning stage. The Adam optimization algorithm is adopted with learning rate set to 1e-3 through all the stages, while different stages leverage different loss items.

The performance of an object detection problem is often affected by extreme class imbalance. For the two-class problem in this paper, specifically, the problem is significant since most pixels are not the desired corners or lines, contributing to most of the loss. Therefore, Focal Loss is used instead of the standard Cross Entropy Loss such that the imbalance problem is solved to some extent:

\begin{equation}
\begin{aligned}
 L_{heatmaps} =
 \begin{cases}
 -(1-p)^2log(p),& \text{if around}(y==1)\\
 -(1-y)^4p^2log(1-p), & \text{otherwise}
 \end{cases}
\end{aligned}
\end{equation}
where p means the predictive value and y means the ground-truth value. The total loss is listed as below:

\begin{equation}
\begin{aligned}
	L_{total} = &L_{heatmap}^{corners} + L_{heatmap}^{entry-lines} + \lambda_{sl} L_{heatmap}^{side-lines} \\
	 &+ \lambda_{\CEN} L_{\CEN} + \lambda_{L1} L_1
\end{aligned},
\end{equation}
where a suppression factor for losses of the less sensitive separating lines is set to $\lambda_{sl} = 0.1$, the regularization term for Contribution Evaluation Networks and the $L1$ regularization for pruning is set to $\lambda_{L1} = 0.05$.

In the pre-training stage, we use 5 epochs for warm-up with only heat-map losses. The model will continue to train 10 epochs with $L_{CEN}$ to $L_{total}$ added, before entering the selecting stage where we determine dilation values of kernels. The decision is made one by one from the first layer to the last and around 2 to 3 training epochs for each layer. In the pruning stage, we remove the $L_{CEN}$, add $L_{1}$ and train the model for at least another 100 epochs. After each epoch, the global contribution of each channel will be calculated according to formula~\eqref{contribution} and we use the pruning strategy described in the \ref{pruning_str} to remove redundant channels. We finally fine-tuned the network by removing both two regularizations and train last 15 epochs to eliminate their effects.

\subsection{Structure Comparison}
To determine the number of hourglasses to be stacked, we first trained both twice stacked hourglasses with intermediate supervision and single hourglass network. According to the experiment results, although multi-hourglass network can improve a bit accuracy and the convergence speed, single-hourglass network is sufficient for the task in this paper. Considering that we want to perform real-time detection on CPU, we finally opt for the single-hourglass network.

\subsection{Selecting Kernels}
Since the network needs to be pruned after selecting, the initial number of channels is set to 64, thereby providing a larger search space.

Different convolution kernel sizes collect information in different receptive fields.
The same receptive field can be achieved by either using standard convolution or using dilated convolution with smaller parameter size. In practice, the kernel with less parameters and computation may or may not run faster than the component.
In this work, when the feature size is less than 28$\times$28, dilated convolution with a fixed kernel size of 3$\times$3 is used for better efficiency, standard convolution is used otherwise because the pytorch implementation (version 1.2.0) of standard convolution is optimized and actually runs faster at larger feature sizes.

\begin{figure}[tbp]
 \centering
 \includegraphics[width=60mm]{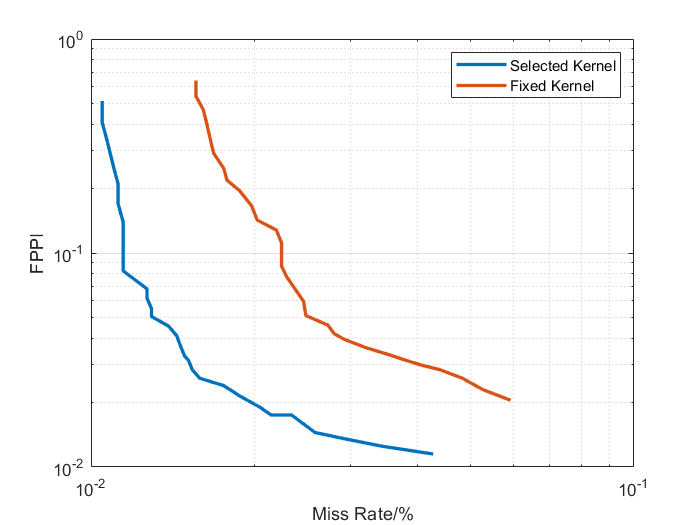}
 \caption{Comparison of selected and fixed kernel networks}
 \label{Different Kernel}
\end{figure}

We trained two networks under the same conditions, except that one using the selected kernel and the other using fixed 3$\times$3 standard convolution. The comparison of their Miss Rate vs. False Positive Per Image profile is shown in Fig.~\ref{Different Kernel}. Obviously, selected receptive fields achieve better performance.

\subsection{Pruning Channels}

\begin{figure}[tbp]
 \centering
 \includegraphics[width=60mm]{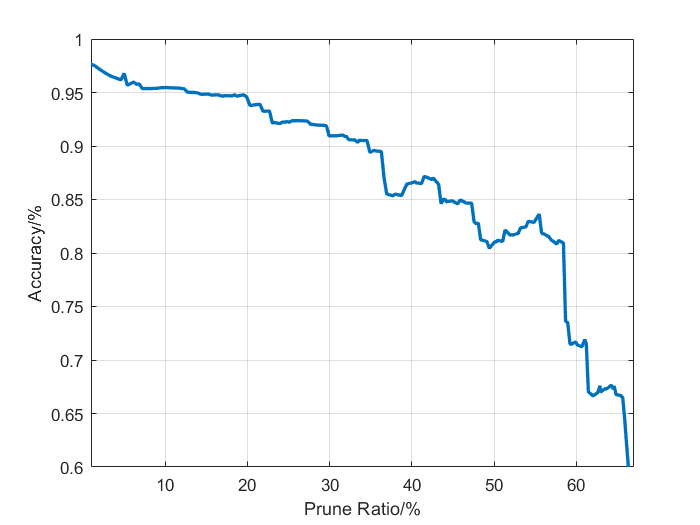}
 \caption{Location Accuracy of Key Points in the pruning process.}
 \label{Accuracy of prune approach}
\end{figure}

Fig.~\ref{Accuracy of prune approach} shows the model accuracy during the learning (training \& pruning) process. Accuracy decreases more and more rapidly when prune ratio increases.

\begin{figure}[tbp]
 \centering
 \includegraphics[width=60mm]{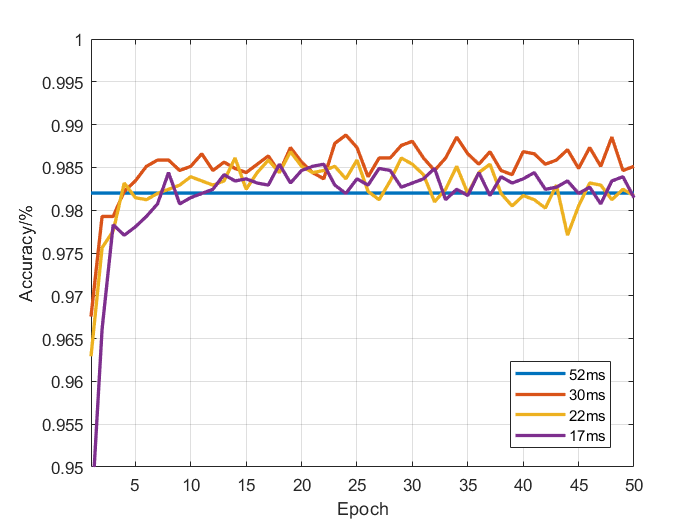}
 \caption{Accuracy in fine-tuning process.}
 \label{Accuracy of fine-tuning}
\end{figure}

To find a balance between network speed and accuracy, we chose three different pruning configurations for fine-tuning, whose inference time is approximately 30 ms, 22 ms and 17 ms respectively. The accuracy of the model without pruning is added as a baseline (with process time of 52 ms). As is shown in Fig.~\ref{Accuracy of fine-tuning}, after 50 epochs of fine-tuning, accuracy of the three pruned networks is generally restored to the level before pruning. It can be noticed that the accuracy of the lighter network goes slightly down in the later training epochs and we have the best parameters at about the 34th epoch.

\subsection{Evaluation and Comparison with Other Methods}
We calculated the localization error of the parking slot markings on the validate set and compared it with the results of other object detection methods, The result is shown in Table~\ref{Localization Error of Different Methods}.


\begin{table}[tbp]
 \caption{Localization Error of Different Methods (other results come from~\cite{zhang_vision-based_2018-1}).}
 \label{Localization Error of Different Methods}
 \begin{center}
 \begin{tabular}{|c||c|}
 \hline
 Method & Localization Error (Unit: cm)\\
 \hline
 Faster R-CNN \cite{ren_faster_2015} & 6.12$\pm$3.87\\
 SSD \cite{liu_ssd:_2016} & 2.52$\pm$1.95\\
 DeepPS (YOLO v2-based)~\cite{zhang_vision-based_2018-1} & 2.58$\pm$1.75\\
 Our Method & 1.51$\pm$2.14\\
 \hline
 \end{tabular}
 \end{center}
\end{table}

For the overall slot detection, the comparison of the results can be seen in Table~\ref{Accuracy of different Methods}. Compared with the previous methods, our method has gained a lot of reduction in the amount of parameters while sacrificing a bit of accuracy, which makes it easier to arrange into limited storage space and computing power such as on-board equipment.


\begin{table}[tbp]
 \caption{Accuracy of different Methods.}
 \label{Accuracy of different Methods}
 \begin{center}
 \begin{tabular}{|c||c||c||c|}
 \hline
 Method & Parameter size/MB & Precision & Recall\\
 \hline
 PSD\_L \cite{zhang_vision-based_2018} (16cm) & 8.38 MB & 98.55\% & 84.64\% \\
 DeepPS \cite{zhang_vision-based_2018-1} (16cm) & 255 MB & \textbf{99.54\%} & \textbf{98.89}\% \\
 Our Method (6cm) & \textbf{2.39 MB} & 98.01\% & 97.31\% \\
 Our Method (16cm) & \textbf{2.39 MB} & 98.26\% & 97.56\% \\
 \hline
 \end{tabular}
 \end{center}
\end{table}

The frame rate of the model with an input image size of 224$\times$224 can reach 30.9FPS on the CPU to meet real-time detection requirements and it reaches about 150FPS on GPU with the post-processing algorithms are still running on the CPU.

\begin{figure}[tbp]
 \centering \includegraphics[width=75mm]{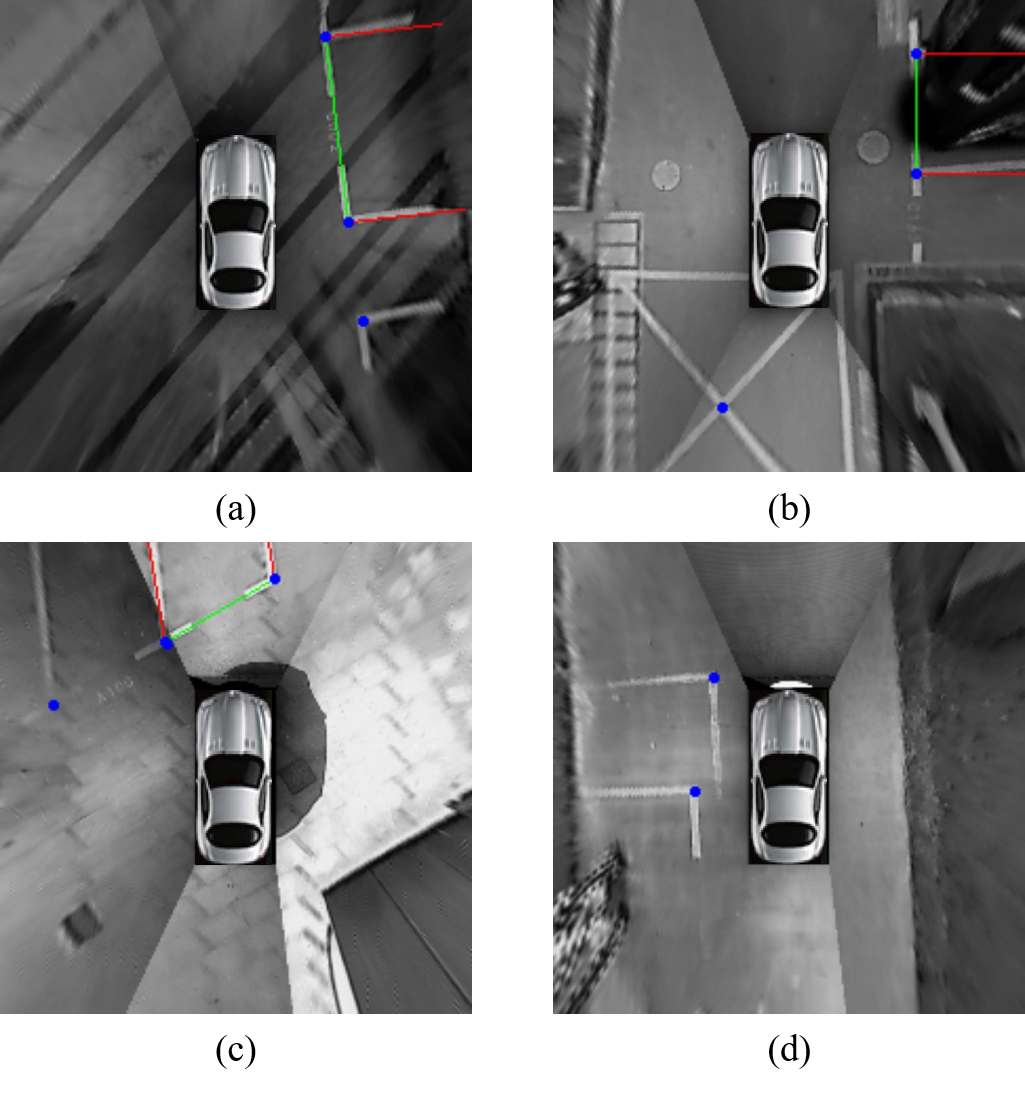}
 \caption{Different results under shadows and interference:
 (a) yields correct corners and slots;
 (b) encounters a misleading ground marking and gives a wrong corner, but excludes it during slot inference;
 (c) yields the correct corners yet misses one slot as a result of the ground marking degradation;
 (d) gets completely wrong corners because of the misleading ground marking, yet successfully excludes the corners as no proper slot can be formed.}
 \label{Masks and Slots}
\end{figure}

Fig.~\ref{Masks and Slots} visualizes four examples of the detection results, in which the corners are marked with blue points. Also the pruned network can recognize parking slots with non-right angled separating lines and entry lines. The model can also identify corners that doesn't belong to any slot, which is shown in the figure but excluded from the slot list during post-processing.

In the above examples, it suggests that by leveraging both point and line detection, the result can be improved under challenging situations. Yet when line detection fails or yields poor results, the parking slot is likely to be excluded from the final result. However, from a practical point of view, here the missed slot detection (false negative) is more acceptable than the false detection (false positive).


\section{CONCLUSIONS}
In this paper, we propose a deep-learning-based model to detect the parking corners and lines for slots inference. The hourglass architecture is used to generate point and line features, and the features are leveraged jointly for parking slot detection. The result is quite promising in accelerating fully convolutional networks without complicate architecture searching methods.

At the same time, the proposed effective network design solution, using the SP module to automatically outline the shape of the network and choose the optimal convolutional kernel configurations during training, can be borrowed for many other applications. The SP module can directly replace the convolution blocks in any network, and get free performance improvement by sacrificing some training speed but without additional inference overhead.


\section*{ACKNOWLEDGMENT}

This work is funded by the Open Project of State Key Laboratory of Advanced Design and Manufacturing for Vehicle Body with funding number 31815005 and by the he National Natural Science Foundation of China with funding number 61906138.


\bibliographystyle{unsrt}
\bibliography{ref}

\end{document}